\documentclass[10pt,twocolumn,letterpaper]{article}

\usepackage{cvpr}
\usepackage{times}
\usepackage{epsfig}
\usepackage{graphicx}
\usepackage{amsmath}
\usepackage{amssymb}

\usepackage{graphicx}
\usepackage{subfigure}
\usepackage{epstopdf}
\usepackage{algorithm}
\usepackage{algorithmic}
\usepackage{verbatim}
\usepackage{marvosym}
\usepackage{hyperref}
\cvprfinalcopy % *** Uncomment this line for the final submission

\def\cvprPaperID{****} % *** Enter the CVPR Paper ID here

\begin{document}

%%%%%%%%% TITLE
\title{Probability Weighted Compact Feature for Domain Adaptive Retrieval}

\author{Fuxiang Huang$^1$, Lei Zhang$^{1(}$\textsuperscript{\Letter}$^)$, Yang Yang$^2$, Xichuan Zhou$^1$\\
$^1$Learning Intelligence \& Vision Essential (LiVE) Group\\
$^1$School of Microelectronics and Communication Engineering, Chongqing University, China\\
$^2$University of Electronic Science and Technology of China\\
{\tt\small \{huangfuxiang, leizhang, zxc\}@cqu.edu.cn,}
%\and Lei Zhang*
%\and
%Yang Yang
%\and
%Xichuan Zhou\\
%University of Electronic Science\\
% and Technology of China\\
{\tt\small dlyyang@gmail.com}
}
\maketitle
%\thispagestyle{empty}

%%%%%%%%% ABSTRACT
\begin{abstract}
Domain adaptive image retrieval includes single-domain retrieval and cross-domain retrieval. Most of the existing image retrieval methods only focus on single-domain retrieval, which assumes that the distributions of retrieval databases and queries are similar. However, in practical application, the discrepancies between retrieval databases often taken in ideal illumination/pose/background/camera conditions and queries usually obtained in uncontrolled conditions are very large. In this paper, considering the practical application, we focus on challenging cross-domain retrieval. To address the problem, we propose an effective method named Probability Weighted Compact Feature Learning (PWCF), which provides inter-domain correlation guidance to promote cross-domain retrieval accuracy and learns a series of compact binary codes to improve the retrieval speed. First, we derive our loss function through the Maximum A Posteriori Estimation (MAP): Bayesian Perspective (BP) induced focal-triplet loss, BP induced quantization loss and BP induced classification loss. Second, we propose a common manifold structure between domains to explore the potential correlation across domains. Considering the original feature representation is biased due to the inter-domain discrepancy, the manifold structure is difficult to be constructed. Therefore, we propose a new feature named Histogram Feature of Neighbors (HFON) from the sample statistics perspective. Extensive experiments on various benchmark databases validate that our method outperforms many state-of-the-art image retrieval methods for domain adaptive image retrieval. The source code is available at \url{https://github.com/fuxianghuang1/PWCF}.
\end{abstract}
\section{Introduction}
The problem of domain adaptive image retrieval including single-domain retrieval and cross-domain retrieval is an important task for many practical applications. Single-domain retrieval refers to a sort of image retrieval problem that the queries and databases are both from the same domain. On the contrary, cross-domain retrieval means the queries and databases permitting to come from different domains, which is more flexible and applicable in real-world applications. In practice, retrieval databases have often taken in ideal illumination/pose/background/camera conditions and queries usually obtained in uncontrolled conditions, which leads to the large discrepancy between the databases and queries. For example, mobile product image search ~\cite{Fitzgibbon2012} aims at identifying a product, or retrieving similar products from the online shopping domain based on a photo captured in unconstrained scenarios by a mobile phone camera.

\begin{figure}
  \centering
 \centerline{\includegraphics[scale=0.28]{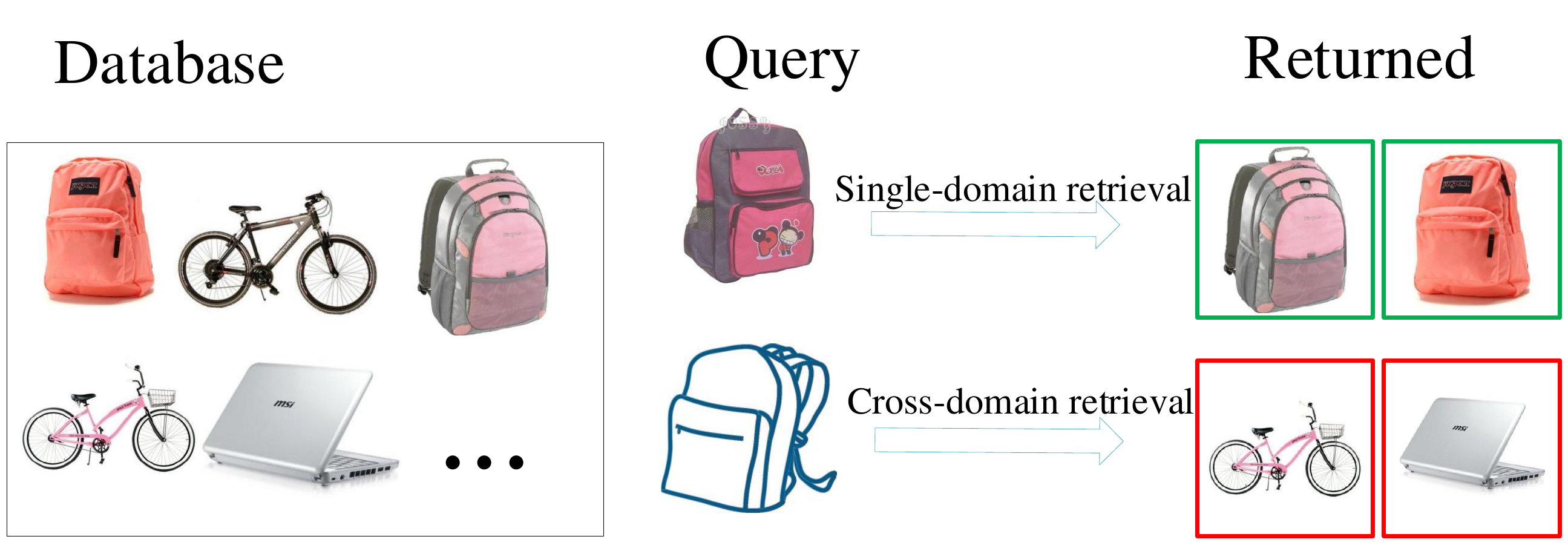}}
  \caption{Illustration of our motivation. Many advanced methods have achieved excellent performance in solving the single-domain retrieval problem, but the performance drops significantly when they are used in cross-domain retrieval. In practice, queries and databases usually come from different domains,  so it is necessary to solve the problem of cross-domain retrieval.}
  \label{fig1}
 \end{figure}

However, as shown in Fig. \ref{fig1}, most of the existing methods only focus on single-domain retrieval and the performance deteriorates rapidly in handling cross-domain retrieval. Few people have proposed solutions to cross-domain retrieval problems. DARN~\cite{JunshiCross} simultaneously integrates the attributes and visual similarity constraint into the retrieval feature learning to solve the cross-domain retrieval problem. However, attributes are usually insufficient and sorting high-dimensional features requires a lot of computation, resulting in slow retrieval.

 \begin{figure*}
  \centering
 \centerline{\includegraphics[scale=0.23]{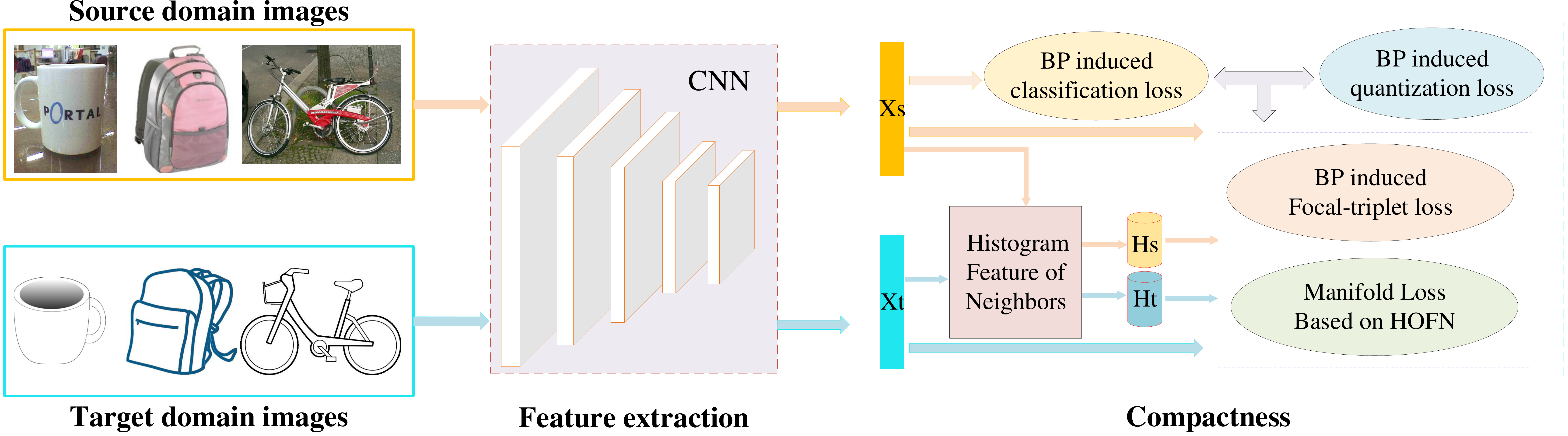}}
  \caption{Diagram of PWCF, which includes four parts: 1) BP induced focal-triplet loss, 2) BP induced classification loss, 3) BP induced quantification loss and 4) manifold loss based on Histogram Feature of Neighbors.}
  \label{fig2}
\end{figure*}
Recently, due to the low storage and high computation efficiency of binary codes, the hashing algorithm has been widely used for many applications~\cite{Gionis1999Similarity, Hao2017Unsupervised, Jiang2013Query, Jinhui2015Neighborhood, Liu2016Query, Norouzi2012Hamming, Tang2015Neighborhood, Wang2013Learning, Zhang2017Semi}. Hashing aims to map high-dimensional content features of samples into Hamming space (binary space) and generate a set of low-dimensional binary codes to represent samples. In consequence, the cost of data storage can be largely reduced, thus the retrieval speed can be improved with the Hamming distance using binary operation (XOR).

However, most of the existing hashing methods~\cite{jin2014density, Kong2012Isotropic, Lin2014Fast, Liu2018Ordinal, NIPS2014_5332, Liu2011Hashing, Strecha2011LDAHash, Weiss2008Spectral} assume that the distributions of retrieval databases and queries are similar while ignoring the inter-domain discrepancy, which makes them difficult to accurately capture correlations between cross-domain samples. Consequently, although most of the existing hashing methods have achieved significant performance for single-domain retrieval, they perform poorly when queries and databases come from different domains.

To address the above problem, we propose an effective domain adaptive image retrieval method named Probability Weighted Compact Feature Learning (PWCT), which takes into account the similarity/dissimilarity relation between the samples from different domains to learn compact binary feature representations. Inspired by transfer learning (TL)~\cite{Pan2010A}, we transfer knowledge across different domains to leverage knowledge between different domains and explore cross-domain sample correlations. Our goal is to use the available labeled data as a source domain to help us learn the projection matrix and get more discriminant binary codes. Instead of simply adding source domain data to expand the training set for better retrieval in the target domain, different from existing transfer hashing methods~\cite{zhou2018transfer, Ji2019Optimal}, we focus on exploring the correlation of samples and data distribution discrepancy between domains to achieve good performance on cross-domain retrieval. To improve the performance of cross-domain retrieval, we propose our loss functions from a Bayesian Perspective. Specifically, we derive our loss functions: BP induced focal-triplet loss, BP induced quantization loss and BP induced classification loss by seeking for the Maximum A Posteriori Estimation (MAP) solution to promote the correlation between samples from different domains in Hamming space, ensure the discrimination of binary codes, and reduce the information error causing by quantification.

Besides, considering that the underlying manifold structure across different domains is extremely helpful to capture meaningful nearest neighbors correlation of different domains, we propose a common manifold to capture the inherent neighborhood structure in the source domain and target domain to further ensure that the correlation of different domains is preserved in Hamming space. However, the similarity between the samples from different domains is difficult to measure in terms of original content features. The same class samples from different domains may not be close, caused by inter-domain discrepancy. To handle such a problem, we consider the distribution characteristics of $k$ nearest neighbors for each sample in respective domains and propose a new statistical feature called Histogram Feature of Neighbors (HFON) from the perspective of sample statistics to reduce the influence of data distribution discrepancy between domains. The main contributions and novelties of this paper are summarized as follows:
\begin{itemize}
\item  In this paper, we propose an effective domain adaptive image retrieval method named Probability Weighted Compact Feature Learning (PWCF) to achieve fast and accurate retrieval. Fig. \ref{fig2} shows the framework of our PWCF. To the best of our knowledge, we are the first to propose a new and practical adaptive cross-domain retrieval problem.
\item We propose loss functions named BP induced focal-triplet loss, BP induced quantization loss, and BP induced classification loss which seeks for the Maximum A Posteriori Estimation (MAP) solution to explore the similarity/dissimilarity between samples from different domains, ensure the discrimination, and reduce the information error.
\item In our PWCF, we propose a Histogram Feature of Neighbors (HFON) from the perspective of statistics to reduce the influence of the domain disparity and a common manifold structure based on HFON to further preserve the correlation between samples from different domains.
\item Extensive experiments on various benchmark databases have been conducted. The experimental results verify that our method outperforms many state-of-the-art image retrieval methods for both cross-domain retrieval and single domain retrieval.
\end{itemize}

\section{Probability Weighted Compact Feature Learning}
\subsection{Notations and Definitions}
Suppose that we have $n_t$ target samples unlabeled
 $\mathbf{X}_t=\left\{\mathbf{x}_{t_i}\right\}_{i=1}^{n_t} \in \mathbb{R}^{d \times{n_t}}$ and $n_s$ source samples labeled  $\mathbf{X}_s=\left\{\mathbf{x}_{s_i}\right\}_{i=1}^{n_s} \in \mathbb{R}^{d \times{n_s}}$. $\mathbf{Y}_s=\left\{\mathbf{y}_{s_i}\right\}_{i=1}^{n_s} \in \mathbb{R}^{c \times{n_s}}$, where $\mathbf{y}_{s_i}\in \mathbb{R}^{c \times{1}}$ is the label vector, of which
the maximum item indicates the assigned class of $\mathbf{x}_{s_i}$. We denote $\mathbf{X}=[\mathbf{X}_t,\mathbf{X}_s]$ and $n=n_t+n_s$. We aim to learn a set of compact binary codes $\mathbf{B}_{t}=\left\{\mathbf{b}_{t_{i}}\right\}_{i=1}^{n_{t}} \in\{-1,1\}^{r \times n_{t}}$ and $\mathbf{B}_{s}=\left\{\mathbf{b}_{s_{i}}\right\}_{i=1}^{n_{s}} \in\{-1,1\}^{r \times n_{s}}$  to represent the samples, where $\mathbf{b}_{t_i}$ is the corresponding binary codes of $\mathbf{x}_{t_i}$ and $\mathbf{b}_{s_i}$ is the corresponding binary codes of $\mathbf{x}_{s_i}$. $d$ and $r$ represent the original content feature dimension of each sample and the length of binary codes, respectively. In PWCF, both the data of the source domain and the target domain are used to learn a projection $\mathbf{W}\in\mathbb{R}^{d \times{r}}$. Then, the $r$-dimensional feature ( i.e., the continuous real values of binary codes) is denoted as $f_{i}=\mathbf{W}^{\top}\mathbf{x}_{i} $. The binary codes are quantified as $\mathbf{b}_{i}=sgn(f_{i})\in\{-1,1\}^{r \times 1}$. Here sgn($v$) is the sign function, which returns $1$ if $v\geq 0$ and $-1$ otherwise. In this paper, $\|\cdot\|$  is the $\ell_{2}$ norm for vectors and Frobenius norm for matrices.

\subsection{Compactness: A Bayesian Perspective}
In order to achieve higher accuracy, we hope to explore the correlation of different samples. Given a triplet $(\mathbf{x}_{i},\mathbf{x}_{j},\mathbf{x}_{k})\in\mathbf{X}$, let $s_{ij}$ represent the pair-wise similarity between $\mathbf{x}_{i}$ and $\mathbf{x}_{j}$. $s_{ij}=1$ means they have the same label. Instead, $s_{ij}=0$ means they have different labels.

Without loss of generality, let $ p\left(f_{i}, f_{j}, f_{k} | s_{ij}, s_{ik}\right)$ be the posterior probability of feature representation  $f_{i}$, $f_{j}$, $f_{k}$ for triplet sample set $\mathbf{x}_{i}$, $\mathbf{x}_{j}$, $\mathbf{x}_{k}$. Here we suppose that $f_{i}$, $f_{j}$, $f_{k}$ are the $r$-dimensional features of samples $\mathbf{x}_{i},\mathbf{x}_{j},\mathbf{x}_{k}$, respectively. With the assumption of conditional independence of each pair and Bayesian formulation, the joint posterior probability density function of the triplet training set can be generally represented as:
\begin{equation}
\begin{aligned}
&\prod_{i,j,k \in\mathbf{X}} p\left(f_{i}, f_{j}, f_{k} | s_{ij}, s_{ik}\right) \Leftrightarrow\\
&\prod_{i,j,k \in\mathbf{X}} p\left(s_{ij}, s_{ik} | f_{i}, f_{j}, f_{k}\right) p\left(f_{i}\right) p\left(f_{j}\right) p\left(f_{k}\right)
\end{aligned}
\label{eq1}
\end{equation}
where $p\left(s_{ij}, s_{ik} | f_{i}, f_{j}, f_{k}\right)$ is the likelihood probability and $p\left(f_{i}\right)$,  $p\left(f_{j}\right)$ and $p\left(f_{k}\right)$ are the prior probability of $r$-dimensional feature. We suppose that the likelihood probability density function to be exponential distribution, considering that the exponential distribution has shown fast convergence to a stable state. Let $d_{ij}=\left\|f_{i}-f_{j}\right\|^{2}$ and $d_{ik}=\left\|f_{i}-f_{k}\right\|^{2}$. Considering the sample pair similarity, the likelihood probability density function is expressed as:
\begin{equation}
\begin{aligned}
&p\left(s_{ij}, s_{ik} | f_{i}, f_{j}, f_{k}\right)=\\
& \left\{
\begin{array}{lcl}
e^{-\left|d_{ij}-d_{ik}+m\right|},&& {if\ s_{ij}=1, s_{ik}=0}\\
e^{-\left|-d_{ij}+d_{ik}+m\right|},&& {if\ s_{ij}=0, s_{ik}=1}\\
e^{-\left|d_{ij}+d_{ik}\right|},&& {if\ s_{ij}=1, s_{ik}=1}\\
e^{-\left|-d_{ij}-d_{ik}\right|},&& {if\ s_{ij}=0, s_{ik}=0}\\
\end{array} \right.\\
&=e^{-\left|(-1)^{\overline{s}_{ij}} d_{ij}+(-1)^{\overline{s}_{ik}} d_{ik}+\alpha \cdot m\right|}
\end{aligned}
\label{eq2}
\end{equation}
where $\overline{s}_{ij}=-{s}_{ij}$, $\alpha=s_{ij} \oplus s_{ik}$ and $\oplus$ is the XOR operation. $m$ is the margin. The purpose of this setting is to make the samples of the same class closer and the samples of different classes farther.

We aim to seek the solution of Maximum A Posteriori Estimation (MAP) of Eq. (\ref{eq1}) from the Bayesian perspective. To mitigate the influence of the likelihood probability of hard pairs on the posterior probability maximization, we add a modulating factor $\omega_{ijk}$ to the likelihood probability where $\omega_{ijk}=\left(1-p\left(s_{ij}, s_{ik} | f_{i}, f_{j}, f_{k}\right)\right)^{\gamma}$ and $\gamma \geq 0$. In other words, the modulating factor reduces the contribution of easy pairs and penalizes more on those hard pairs. For convenience, the same (different) labeled samples which have a large (small) distance are named as hard pairs, and the same (different) labeled samples that have a small (large) distance are named as easy pairs.

Besides, considering the quantization loss and discrimination of binary code, the prior probability is written as
$p\left(f_{i}\right)=e^{-\theta d\left(\mathbf{b}_{i},f_{i}\right)}\cdot e^{-\lambda_1 d(\mathbf{y}_i,\mathbf{C}^{\top}\mathbf{b}_i)}\cdot e^{-\lambda_2{\|\mathbf{C}\|}^2}$, where $\theta$, $\lambda_1$ and $\lambda_2$ are hyper-parameters. $\mathbf{C}$ is a classifier and we will discuss the detail later in classification loss. By taking the natural logarithm, our objective function is written as:
\begin{equation}
\begin{aligned}
\max &\sum_{i,j,k \in\mathbf{X}} \omega_{ijk} \log p\left(s_{ij}, s_{ik} | f_{i}, f_{j}, f_{k}\right)+\\
&\sum_{i \in\mathbf{X}} \log p\left(f_{i}\right)+\sum_{j \in\mathbf{X}} \log p\left(f_{j}\right)+\sum_{k \in\mathbf{X}} \log p\left(f_{k}\right)
\end{aligned}
\label{eq3}
\end{equation}
In optimization, we consider the case that positive pairs and negative pairs both exist. In order to construct triplets, we can set $\mathbf{x}_i$ as the anchor, $\mathbf{x}_j$ is similar to the anchor, and $\mathbf{x}_k$ is dissimilar to the anchor. Then Eq. (\ref{eq3}) is
\begin{equation}
\begin{aligned}
\min&\sum_{i,j,k\in\mathbf{X}} (1-e^{[d_{ij}-d_{ik}+m]_+})[d_{ij}-d_{ik}+m]_+\\
&+\theta\sum_{i \in\mathbf{X}}d(\mathbf{b}_i,f_i)+\lambda_1\sum_{i \in\mathbf{X}}d(y_i,\mathbf{C}^{\top} \mathbf{b}_{i})+\lambda_2\|\mathbf{C}\|^{2}
\end{aligned}
\label{eq4}
\end{equation}
where $[x]_+$ denotes the operator of $\max(x,0)$, which makes sure $p\left(s_{ij}, s_{ik} | f_{i}, f_{j}, f_{k}\right)\in (0,1]$ and improves the convergence. Clearly, if without $\max(x,0)$, the probability of the exponential probability in Eq.(\ref{eq2}) may be larger than 1. So, in Eq.(\ref{eq4}), the $\max(x,0)$ is naturally resulted with clear probabilistic interpretation.

\begin{figure}
  \centering
 \centerline{\includegraphics[scale=0.5]{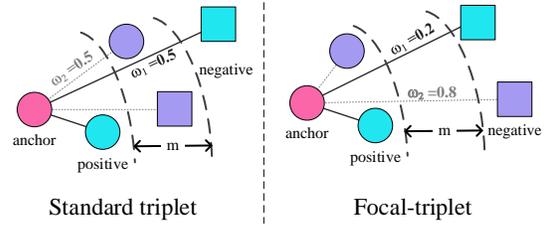}}
  \caption{Illustration of the proposed BP induced focal-triplet loss: Standard triplet makes the positive sample closer to the anchor and the negative sample further away from the anchor by the same force (weight). However, positive samples and negative samples may not be separated for hard pairs in this case, which results in training instability. To address it, our BP induced focal-triplet loss can down-weight easy pairs and up-weight hard pairs so as to the distance to the cross-domain positives can be minimized and the distance to the cross-domain negatives can be maximized.}
  \label{fig3}
\end{figure}

\textbf{BP induced Focal-triplet loss.}
 The first term in Eq. (\ref{eq3}) is a variant of standard triplet loss named BP induced focal-triplet loss. If we enumerate all the sample pairs, it will take a lot of time for training. So we will just pick some cross-domain triplets, which are more effective at promoting inter-domain correlations. In other words, we construct cross-domain triplets before training. Specifically, for each sample, if it comes from the source domain, we select a positive sample and a negative sample from the target domain. Otherwise, if it is from the target domain, we select a positive sample and a negative sample from the source domain. Since there is no label in the target domain, we first use the source domain data to predict the pseudo-label of the target domain by the KNN algorithm. We can get $n$ cross-domain triplets. For ease of understanding, let's $(\mathbf{x}_{i}^{g},\mathbf{x}_{i,p}^{\overline{g}},\mathbf{x}_{i,n}^{\overline{g}}),i\in [1,n]$ represent all selected triplets, where $g$  and $\overline{g}$ come from different domain. If $g\in \mathbf{X}_s$, then $\overline{g} \in\mathbf{X}_t$. Otherwise, if $g\in \mathbf{X}_t$, then $\overline{g} \in\mathbf{X}_s$. Then, the BP induced focal-triplet loss can be written as:
\begin{equation}
\begin{aligned}
&\mathcal{T}ri = \sum_{i,j,k\in\mathbf{X}} (1-e^{[d_{ij}-d_{ik}+m]_+})[d_{ij}-d_{ik}+m]_+=\\
&\sum_{i=1}^{N}\!\omega_i\!\left[\left\|\mathbf{W}^{\top}\mathbf{x}_{i}^{g}\!-\!\mathbf{W}^{\top} \mathbf{x}_{i,p}^{\overline{g}}\right\|^{2}\!\!\!\!-\!\!\left\|\mathbf{W}^{\top}\mathbf{x}_{i}^{g}\!-\!\mathbf{W}^{\top} \mathbf{x}_{i,n}^{\overline{g}}\right\|^{2}\!\!\!\!+\!m\right]_+
\end{aligned}
\label{eq5}
\end{equation}
where $\omega_i$ is the weight of $i^{th}$ group selected triplet and $\omega_i=\left(1-e^{-\left(\left\|\mathbf{W}^{\top}\mathbf{x}_{i}^{g}-\mathbf{W}^{\top} \mathbf{x}_{i,p}^{\overline{g}}\right\|^{2}-\left\|\mathbf{W}^{\top}\mathbf{x}_{i}^{g}-\mathbf{W}^{\top} \mathbf{x}_{i,n}^{\overline{g}}\right\|^{2}+m\right)}\right)^\gamma$. As shown in Fig. \ref{fig3}, the focal-triplet loss, which is a variant of standard triplet loss, imposes different importance for different triplets by down-weighting easy pairs and up-weighting hard pairs. In the training phase, we choose the hard triplets that satisfy the maximization of intra-class distance and the minimization of inter-class distance to improve the training speed. Considering the data distribution discrepancy in different domains, the Euclidean distance of the original content feature extracted from different domains may not measure the similarity of samples. So we use the Histogram Feature of Neighbors rather than the original content feature to calculate the distance of different samples across domains. The Histogram Feature of Neighbors will be explained in detail in the next section.

\textbf{BP induced quantization loss.}
The second term in Eq. (\ref{eq4}) is named BP induced quantization loss, which aims to reduce quantization error between binary codes and low-dimensional feature representation obtained by mapping (i.e., the continuous real values of binary codes). The BP induced quantization loss can be formulated as:
\begin{equation}
\begin{aligned}
\mathcal{Q}&=\sum_{i \in\mathbf{X}}d(\mathbf{b}_i,f_i)=\sum_{i=1}^{n}\left\|\mathbf{b}_i- \mathbf{W}^{\top} \mathbf{x}_{i}\right\|^{2}\\
&=\left\|\mathbf{B}_t- \mathbf{W}^{\top} \mathbf{X}_t \right\|^{2}+\left\|\mathbf{B}_s- \mathbf{W}^{\top} \mathbf{X}_s \right\|^{2}
\end{aligned}
\label{eq6}
\end{equation}

\textbf{BP induced classification loss.}
The third term in Eq. (\ref{eq4}) is named BP induced classification loss. Inspired by SDH~\cite{shen2015supervised}, we consider that good binary codes should be with good discrimination. We take advantage of the label information to train a classifier $\mathbf{C}$ and $\mathbf{C}^{\top}\mathbf{b}_i$ represent the predicted label of the $i^{th}$ sample. We want to use binary codes to predict labels that are as authentic as possible. In this paper, to avoid the negative impact of pseudo labels, we only use the source domain sample when we calculate the classification loss. The BP induced classification loss can be formulated as:
\begin{equation}
\begin{aligned}
\mathcal{C}&=\sum_{i \in\mathbf{X}}d(y_i,\mathbf{C}^{\top} \mathbf{b}_{i})=\sum_{i=1}^{n}\left\|\mathbf{y}_{i}-\mathbf{C}^{\top} \mathbf{b}_{i}\right\|^{2}\\
&\approx\left\|\mathbf{Y}_{s}-\mathbf{C}^{\top} \mathbf{B}_{s}\right\|^{2}
\label{eq7}
\end{aligned}
\end{equation}
The regularization $\|\mathbf{C}\|^{2}$, i.e., the last term in Eq. (\ref{eq3}), is used to avoid trivial solution and overfitting.
\subsection{Manifold Loss Based on Histogram Feature of Neighbors}
\begin{figure}
  \centering
 \centerline{\includegraphics[scale=0.4]{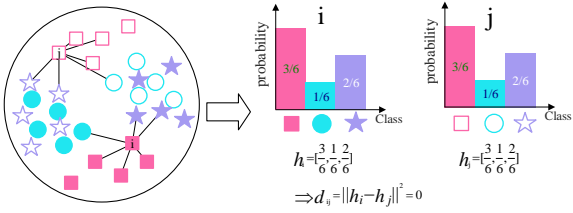}}
  \caption{Illustration of the proposed Histogram Feature of Neighbors (HFON): We use different shapes to represent samples of different classes, The samples from the source domain and target domain are represented by solid and hollow shapes, respectively. $i$ and $j$ belong to the same class, but the distance is far. To measure the similarity of cross-domain samples more accurately, we proposed a Histogram Feature of Neighbors according to the neighbor relationship in a domain. For example, we find 6 nearest neighbors of $i$ and $j$ in their respective domains and calculate the probability of each class of these nearest-neighbor samples. HFON vector is made up of these probabilities. Then the HFON distances of similar samples in different domains are close.}
  \label{fig4}
\end{figure}

We argue that the nearest-neighbor relationship of samples in a single domain is regular. In other words, if two samples from different domains are similar, the classes of their neighbors in their respective domains should be similar. Based on this assumption, we propose a statistical feature named Histogram Feature of Neighbors (HFON) to reduce the domain disparity. Specifically, we use $\mathbf{h}_{i}\in \mathbb{R}^{c \times{1}}$ to represent the HFON vector of $\mathbf{x}_{i}$ and $c$ is the number of classes. We find $k$ nearest neighbors of each sample in their respective domains and calculate the probability of each class of these nearest-neighbor samples. The $a^{th}$ element of the HFON can be written as  $\mathbf{h}_{i_a}=\frac{n_{i_a}}{k}$  where $a\in[1,c]$, $n_{i_a}$ represents the total number of samples belonging to class $c$ in the $k$ nearest neighbors of the $i^{th}$ sample. Fig. \ref{fig4} shows the details of the Histogram Feature of Neighbors.

The underlying manifold structure across different domains, which is extremely helpful to capture meaningful nearest neighbors of different domains. Therefore, we want to keep the common manifold structure by taking advantage of local similarities. To minimize the representation error of the low-dimensional features between different neighbor samples, the manifold  loss can be written as:
\begin{equation}
\begin{aligned}
\mathcal{M}&=\sum_{i,j \in\mathbf{X}} \|f_i-f_j\|^{2}Z_{ij}\\
&=\sum_{i=1}^n\sum_{j=1}^n\left\|\mathbf{W}^{\top} \mathbf{x}_{i}-\mathbf{W}^{\top} \mathbf{x}_{j}\right\|^{2} Z_{i j}\\
&=\mathbf{W}^{\top} \mathbf{X} \mathbf{L}\mathbf{X}^{\top} \mathbf{W}
\label{eq8}
\end{aligned}
\end{equation}
Similar to LPP~\cite{He2004Locality}, $\mathbf{L}=\mathbf{D}-\mathbf{Z}$ is the Laplacian matrix and $D_{ii}=\sum_{j} {Z}_{ij}=\sum_{i} {Z}_{ij}$. Here $\mathbf{Z}$ is a sparse symmetric $n\times n$ matrix with $Z_{ij}$ having the weight of the
edge connecting $\mathbf{x}_i$ and $\mathbf{x}_j$, and 0 if there is no such connection. To reduce the domain disparity, let $Z_{ij}=e^{-\frac{\left\|\mathbf{x}_{i}-\mathbf{x}_{j}\right\|^{2}}{\sigma^2}}$ when $\mathbf{x}_i$ and $\mathbf{x}_j$ come from the same domain. Otherwise, $Z_{ij}=e^{-\frac{\left\|\mathbf{h}_{i}-\mathbf{h}_{j}\right\|^{2}}{\sigma^2}}$ when $\mathbf{x}_i$ and $\mathbf{x}_j$ come from different domains where $\mathbf{h}_{i}$ and $\mathbf{h}_{j}$ denote the HFON with respect to $\mathbf{x}_{i}$ and $\mathbf{x}_{j}$, respectively.

\textbf{Overall objective function.} Finally, the overall objective function is rewritten as:
\begin{equation}
\begin{aligned}
&\min_{\mathbf {W},\mathbf{C},\mathbf{B}_t,\mathbf{B}_s} \mathcal{T}ri+\theta\mathcal{Q}+\lambda_1\mathcal{C}+\lambda_2\|\mathbf{C}\|^{2}+\lambda_3\mathcal{M}\\
&s.t.\mathbf{W}^{\top} \!\mathbf{W}=\mathbf{I},  \mathbf{B}_t\!=sgn(\mathbf{W}^{\top}\! \mathbf{X}_t), \mathbf{B}_{s}\!=sgn(\mathbf{W}^{\top} \!\mathbf{X}_{s})
\end{aligned}
\label{eq9}
\end{equation}
where the constraint, $\mathbf{W}^{\top} \mathbf{W}=\mathbf{I}$, is used to make $\mathbf{W}$ be orthogonal projections in order to guarantee the discrimination of binary codes.
\subsection{Optimization}
In this paper, we adopt an alternating optimization procedure to iteratively optimize $\mathbf {W}$, $\mathbf{C}$, $\mathbf{B}_t$ and $\mathbf{B}_s$. As the non-convex sgn($\cdot$) function makes Eq. (\ref{eq8}) a NP-hard problem, we relax the sgn($x$) function as its signed magnitude $x$~\cite{Ji2019Optimal}.

\textbf{$\mathbf{W}$-Step.} Given $\mathbf{C}$, $\mathbf{B}_t$ and $\mathbf{B}_s$, updating $\mathbf{W}$ is a typical optimization problem with orthogonality constraints. Let $\mathbf{G}$ be the partial derivative of the objective function Eq. (\ref{eq9}) with respect to $\mathbf{W}$ and $\mathbf{G}$ is represented as:
\begin{equation}
\begin{aligned}
\mathbf{G}\!\!=&2\!\!\sum_{i\in\mathcal{T}^+}\!\!\omega_i((\mathbf{x}_{i}^{g}\!\!-\!\!\mathbf{x}_{i,p}^{\overline{g}})
(\mathbf{x}_{i}^{g}\!\!-\!\!\mathbf{x}_{i,p}^{\overline{g}})\!^{\top}\!\!\!-\!\!(\mathbf{x}_{i}^{g}\!\!-\!\!\mathbf{x}_{i,n}^{\overline{g}})
(\mathbf{x}_{i}^{g}\!\!-\!\!\mathbf{x}_{i,n}^{\overline{g}})\!^{\top}\!\!)\!\mathbf{W}\\
&+2\theta(\mathbf{X}_t\mathbf{X}_t^{\top}\mathbf{W}-\mathbf{X}_t\mathbf{B}_t^{\top}+\mathbf{X}_s\mathbf{X}_s^{\top}\mathbf{W}-\mathbf{X}_s\mathbf{B}_s^{\top})\\
&+2\lambda_3 \mathbf{X}\mathbf{L}\mathbf{X}^{\top}\mathbf{W}
\end{aligned}
\label{10}
\end{equation}
where $\mathcal{T}^+$ contains all selected cross-domain triplets and
 $\left\|\mathbf{W}^{\top}\mathbf{x}_{i}^{g}-\mathbf{W}^{\top} \mathbf{x}_{i,p}^{\overline{g}}\right\|^{2}-\left\|\mathbf{W}^{\top}\mathbf{x}_{i}^{g}-\mathbf{W}^{\top} \mathbf{x}_{i,n}^{\overline{g}}\right\|^{2}+m\geq0$. Based on the orthogonal constraint optimization procedure in ~\cite{Wen2013A}, we can define a skew-symmetric matrix~\cite{Armstrong2005Numerical} as $\mathbf{A}=\mathbf{G}\mathbf{W}^{\top}-\mathbf{W}\mathbf{G}^{\top}$. Then, we adopt Crank Nicolson like scheme to update the orthogonal matrix $\mathbf{W}$
\begin{equation}
\mathbf{W}^{(t+1)}=\mathbf{W}^{(t)}-\frac{\tau}{2} \mathbf{A}\left(\mathbf{W}^{(t)}+\mathbf{W}^{(t+1)}\right)
\label{eq11}
\end{equation}
where $\tau$ denotes the step size. We empirically set $\tau=0.1$. By solving Eq. (\ref{eq11}), we can get
\begin{equation}
\mathbf{W}^{(t+1)}=\mathbf{Q}\mathbf{W}^{(t)}
\label{eq12}
\end{equation}
and $\mathbf{Q}=(\mathbf{I}+\frac{\tau}{2}\mathbf{A})^{-1}(\mathbf{I}-\frac{\tau}{2}\mathbf{A})$.
We iteratively update $\mathbf{W}$ several times based on Eq. (\ref{eq12}) with the Barzilai-Borwein (BB) method~\cite{Wen2013A}. In addition, please note that when iteratively optimizing $\mathbf{W}$, the initial $\mathbf{W}$ is set to be the updated one in the last round. For the first round, $\mathbf{W}$ is initialized by PCA.

\textbf{$\mathbf{C}$-Step.} Given $\mathbf{W}$, $\mathbf{B}_t$ and $\mathbf{B}_s$,
taking the partial deviation of the objective function with respect to $\mathbf{C}$ to be zero, we derive
\begin{equation}
\mathbf{C}=\left(\lambda_1\mathbf{B}_s\mathbf{B}_s^{\top}+\lambda_2 \mathbf{I}\right)^{-1} \lambda_1\mathbf{B}_s \mathbf{Y}_s^{\top}
\label{eq13}
\end{equation}

\textbf{$\mathbf{B}_t$-Step.} Given $\mathbf{W}$, $\mathbf{C}$ and $\mathbf{B}_s$, by relaxing the sign function $sgn(.)$ , the solution can be obtained
\begin{equation}
\mathbf{B}_t=sgn(\mathbf{W}^{\top} \mathbf{X}_t)
\label{eq14}
\end{equation}

\textbf{$\mathbf{B}_s$-Step.} Given $\mathbf{W}$, $\mathbf{C}$ and $\mathbf{B}_t$, we obtain the approximate solution for hash codes by relaxing the sign function.
\begin{equation}
\mathbf{B}_s=sgn\left((\theta\mathbf{I}+\lambda_1\mathbf{C}\mathbf{C}^{\top})^{-1}
(\theta\mathbf{W}^{\top}\mathbf{X}_s+\lambda_1\mathbf{C}\mathbf{Y}_s)\right)
\label{eq15}
\end{equation}
The details of the proposed algorithm are described in Algorithm \ref{Algorithm1}.
\begin{algorithm}
\renewcommand{\algorithmicrequire}{\textbf{Input:}}
\renewcommand{\algorithmicensure}{\textbf{Output:}}
\caption{\quad \textbf{PWCF learning}}
\label{Algorithm1}
\begin{algorithmic}[1]
\REQUIRE Training samples $\mathbf{X}=[\mathbf{X}_t,\mathbf{X}_s]$ and source labels $\mathbf{Y}_s$; code length $r$; maximum iteration times $T$; number of neighbors $k$; parameters $\theta$, $\lambda_1$,  $\lambda_2$ and $\lambda_3$
\ENSURE  $\mathbf {W}$, $\mathbf{C}$, $\mathbf{B}_t$ and $\mathbf{B}_s$
\STATE Obtain the pseudo-labels of target domain by KNN trained with ($\mathbf{X}_s$,$\mathbf{Y}_s$).\\
\STATE Calculate the HFON of target domain and source domain, respectively.\\
\STATE Construct cross-domain triples: $(\mathbf{x}_{i}^{g},\mathbf{x}_{i,p}^{\overline{g}},\mathbf{x}_{i,n}^{\overline{g}}),i\in [1,N]$.\\
\STATE  Initialize $\mathbf{W}$ as the top $r$ eigenvectors of $\mathbf{X}\mathbf{X}^{\top}$ by PCA. $\mathbf{B}_t$ and $\mathbf{B}_s$ are random binary matrices, respectively. \\
\STATE Loop until converge or reach maximum iterations:\\
   \textbf{$\mathbf{W}$-Step.}   update $\mathbf{W}$ by solving Eq. (\ref{eq12});\\
   \textbf{$\mathbf{C}$-Step. }    update $\mathbf{C} $  by solving Eq. (\ref{eq13});\\
   \textbf{$\mathbf{B}_t$-Step.} update $\mathbf{B}_t$ by solving Eq. (\ref{eq14});\\
   \textbf{$\mathbf{B}_s$-Step.} update $\mathbf{B}_s$ by solving Eq. (\ref{eq15}).\\
\end{algorithmic}
\end{algorithm}

\textbf{Computation Complexity:} Since the hard triplets and the Laplacian matrix can be pre-computed, the total computation cost of our PWCF in Algorithm 1 is $\mathcal{O}$$( T(n(d^2r+dr)+3d^2r+n_sr^2c+n_tdr+n_s(r^2c^2+r^2cd)))$ and linear to the number of samples, where $n=n_s+n_t$. In practice, $T$, $d$, $r$ and $c$ will be much less than $n$. Hence, the binary codes learning is efficient.

\begin{table*}
\caption{The MAP scores (\%) on MNIST\&USPS, VOC2007\&Caltech101, and Caltech256\&ImageNet databases with varying code length from 16 to 128 for cross-domain retrieval.}
%\small
\resizebox{\textwidth}{22.8mm}{
\begin{tabular}{c|cccccc|cccccc|cccccc}
\hline
        &  \multicolumn{6}{c|}{MNIST\&USPS}  & \multicolumn{6}{c|}{VOC2007\&Caltech101}      & \multicolumn{6}{c}{Caltech256\&ImageNet}\\
\hline
 Bit    &16&32&48&64&96&128     &16&32&48&64&96&128      &16&32&48&64&96&128       \\
\hline
 NoTL   &28.13&30.05&28.24&30.34&31.76&31.72 &35.95&37.86&38.28&38.49&38.67&38.97
 &15.10&19.77&22.80&24.39&26.07&27.28  \\
 SH     &15.71&13.85&12.05&11.78&11.38&11.78 &29.94&30.26&32.51&33.76&32.59&33.03
&10.37&11.67&12.17&11.88&12.67&12.89   \\
 ITQ    &27.38&30.92&31.44&32.25&33.12&33.44 &40.13&39.63&39.45&39.98&39.27&39.89
&16.94&22.00&24.44&26.21&27.96&28.89   \\
 DSH    &21.15&27.53&29.71&26.13&26.60&28.94 &40.97&42.03&43.06&45.81&43.78&42.86
&8.27& 9.60 &11.55&12.34&13.56&15.64    \\
 LSH    &16.25&16.99&23.23&20.38&19.70&26.98 &33.40&33.99&34.03&32.89&34.12&34.50
&5.36&6.72&10.39&12.71&15.60&17.08       \\
 SGH    &24.83&24.78&25.85&27.78&28.26&29.35 &35.77&34.06&33.60&33.11&32.75&32.41
 &12.49&17.23&20.34&21.75&24.46&25.42    \\
 OCH    &18.94&25.73&26.73&26.34&27.88&29.22 &71.50&72.27&72.65&72.71&69.17&68.91
 &11.56&15.36&17.49&20.18&22.00&22.90      \\
 ITQ+ &20.27&20.53&16.77&15.87&17.79&14.90   &35.35&34.48&34.33&34.42&34.05&34.74
 &--&--&--&--&--&--\\
 LapITQ+&26.38&26.31&24.91&24.61&22.04&21.33 &38.95&38.43&39.64&39.35&39.33&38.76
 &--&--&--&--&--&--\\
 GTH    &19.10&24.17&24.27&24.38&23.64&29.36 &36.70&38.95&37.23&37.87&37.70&38.36
 &11.56&14.79&16.97&19.53&20.88&22.38  \\
 KSH    &43.75&46.91&50.02&47.43&45.25&46.81 &74.74&76.05&76.71&76.70&76.22&73.14
  &20.34&12.07&26.77&32.83&35.28&34.49 \\
 SDH    &29.98&43.02&42.57&46.56&42.40&48.12 &67.60&65.75&68.58&65.06&65.66&67.03
 &18.05&25.71&26.23&26.38&26.77&26.29      \\
  \hline
 PWCF  & \textbf{47.47}& \textbf{51.99}&\textbf{51.44} &\textbf{51.75}&\textbf{50.89}&\textbf{53.95}
 & \textbf{79.38}& \textbf{80.42}&\textbf{79.24} &\textbf{79.31}&\textbf{78.15}&\textbf{78.87}
& \textbf{22.46}& \textbf{30.58}&\textbf{35.29} &\textbf{35.24}&\textbf{38.92}&\textbf{40.32}\\
\hline
\end{tabular}}
\label{tab1}
\end{table*}
\begin{table*}
\caption{The MAP scores (\%) on MNIST\&USPS, VOC2007\&Caltech101, and Caltech256\&ImageNet databases with varying code length from 16 to 128 for single-domain retrieval.}
%\small
\resizebox{\textwidth}{22.8mm}{
\begin{tabular}{c|cccccc|cccccc|cccccc}
\hline
        &  \multicolumn{6}{c|}{MNIST\&USPS}  & \multicolumn{6}{c|}{VOC2007\&Caltech101}      & \multicolumn{6}{c}{Caltech256\&ImageNet}\\
\hline
 Bit    &16&32&48&64&96&128     &16&32&48&64&96&128      &16&32&48&64&96&128       \\
\hline
 NoTL   &67.22&69.31&70.52&70.78&71.64&71.88 &98.12&98.13&98.36&98.36&98.58&98.82
  &15.12&20.50&22.66&24.22&25.954&27.43 \\
 SH     &47.07&49.19&49.24&49.64&49.69&49.03 &66.56&64.44&66.39&68.39&67.29&66.12
 &10.44&11.33&12.28&12.24&12.85&13.51 \\
 ITQ    &63.37&69.96&69.53&70.19&71.22&71.59 &99.03&99.15&99.09&99.10&99.14&99.19
 &15.96&20.57&23.07&24.16&26.39&27.63 \\
 DSH    &45.76&54.23&58.03&59.92&61.80&63.50 &94.58&93.88&95.93&97.07&97.93&97.86
&8.09&10.21&11.83&12.89&15.43&15.15  \\
 LSH    &47.21&55.63&59.81&60.54&60.81&62.90 &61.17&65.69&79.25&84.76&88.81&88.19
&5.15&6.87&10.22&12.63&15.37&17.38    \\
 SGH    &58.41&63.61&64.69&65.79&66.55&66.81 &86.06&86.70&88.49&88.68&91.18&91.71
 &12.37&16.75&19.54&20.90&23.37&24.93 \\
 OCH    &53.56&58.49&60.48&63.96&66.29&65.39 &89.36&97.47&98.30&98.45&98.71&99.21
 &10.11&15.05&17.47&19.58&20.70&22.31  \\
 ITQ+  &41.87&37.94&37.00&37.23&35.19&34.88  &64.15&59.00&56.94&56.12&54.74&52.61
 &--&--&--&--&--&-- \\
 LapITQ+&54.21&55.64&53.66&52.58&51.56&49.80 &70.53&69.08&67.95&66.32&69.49&67.59
 &--&--&--&--&--&--\\
 GTH    &53.20&58.78&62.17&63.23&62.94&60.59 &90.25&82.72&92.93&93.73&94.33&87.44
 &11.73&15.02&17.67&19.21&20.87&21.90 \\
 KSH    &26.06&37.11&42.57&41.89&40.89&38.07 &96.98&91.90&93.03&92.58&96.27&95.86
 &16.20&11.23&19.88&25.64&28.07&29.58       \\
 SDH    &50.32&54.20&57.29&57.48&60.64&60.76 &88.49&86.54&88.27&87.28&89.11&89.24
 &13.92&18.72&21.20&21.64&22.74&23.94       \\

  \hline
 PWCF  & \textbf{69.37}& \textbf{70.70}&\textbf{70.94} &\textbf{71.64}&\textbf{73.51}&\textbf{73.89}
 & \textbf{99.67}& \textbf{99.61}&\textbf{99.77} &\textbf{99.66}&\textbf{99.33}&\textbf{99.58}
 & \textbf{21.96}& \textbf{26.56}&\textbf{28.75} &\textbf{29.91}&\textbf{32.92}&\textbf{34.96}

 \\
\hline
\end{tabular}}
\label{tab2}
\end{table*}
\section{Experiment}
\subsection{Experimental Settings}
\textbf{Datasets:} We perform the experiments on four groups benchmark datasets:
\begin{itemize}
\item The \textbf{MNIST}~\cite{L1998Gradient} and USPS~\cite{Hull2002Database} are two famous digital datasets sharing ten handwritten digits from 0 to 9. According to ~\cite{Long2014Transfer}, each image is resized to $16 \times 16$. We use the MNIST as the source domain and USPS is used as the target domain in the \emph{MNIST\&USPS} dataset.
\item  The \textbf{VLCS}~\cite{Torralba2011Unbiased} dataset aggregates photos from Caltech101, LabelMe, Pascal VOC2007 and SUN09, which provides a 5-way multi-class benchmark on five common classes: bird, car, chair, dog and person. In our experiments, every image is represented by a 4096-d CNN feature vector~\cite{Donahue2013DeCAF}. We use VOC2007 dataset including 3376 images as the source domain and Caltech101 dataset containing 1415 images is used as the target domain in the \emph{VOC2007\&Caltech101} dataset.
\item The \textbf{Cross-dataset Testbed}~\cite{Daum2009Frustratingly} is a Decaf7 based cross-dataset image classification dataset, which contains 40 categories of images from 3 domains: 3,847 images in Caltech256, 4,000 images in ImageNet, and 2,626 images for SUN. In our experiments, each image is represented by a 4096-d CNN feature vector~\cite{Donahue2013DeCAF}. Caltech256 is used as the source domain and ImageNet is used as the target domain in the \emph{Caltech256\&ImageNet} dataset.
\item The \textbf{Office-Home} dataset \cite{venkateswara2017Deep} consists of images from 4 different domains: Artistic images (i.e., paintings, sketches and/or artistic depictions), Clip Art images, Product images without background and Real-World images (i.e., regular images captured with a camera). For each domain, the dataset contains images of 65 object categories found typically in Office and Home settings. In our experiments, each image is represented as a 4096-d feature by VGG-16. Each domain is used as the source domain and the target domain, respectively.
\end{itemize}

\textbf{Implementation details:} We choose eleven state-of-the-art hashing methods, including SH ~\cite{Weiss2008Spectral}, ITQ~\cite{Gong2013Iterative}, DSH~\cite{jin2014density}, LSH~\cite{Datar2004Locality}, SGH~\cite{Jiang2015Scalable}, OCH~\cite{Liu2018Ordinal}, GTH~\cite{Ji2019Optimal}, ITQ+~\cite{zhou2018transfer}, LapITQ+~\cite{zhou2018transfer}, KSH~\cite{liu2012supervised} and SDH~\cite{shen2015supervised} as baselines. We use the public codes and suggested parameters of these methods from the corresponding authors. For our PWCF, we empirically set $\theta$ to 1e2, $\lambda_1$ to 1, $\lambda_2$ to 1e3 and $\lambda_3$ to 1e4. For unsupervised methods, we use all the training samples including source domain and target domain in the training phase. For a fair comparison, we introduce a NoTL method that only uses the target domain to train the ITQ model. For supervised methods, we use the training samples and labels of source domain to train. All of the methods use identical training sets and testing sets. Specifically, for each dataset, we randomly select 500 images of the target domain images as a testing set (queries) and rest images as a training set. In the testing phase, class labels are used to determine whether a sample returned for a given query is considered a true positive. Moreover, the widely used criterion, i.e., mean average precision (MAP), is used as the performance metric. To remove the randomness for sampling, we repeat each algorithm 10 times and report their mean of MAP. We also show the precision and recall curves.

To verify the performance of our method in the scenarios of domain adaptive retrieval, we report the retrieval performance including cross-domain retrieval and single-domain retrieval on \emph{MNIST\&USPS}, \emph{VOC2007\&Caltech101}, and \emph{Caltech256\&ImageNet} databases when the code length is set as 16, 32, 48, 64, 96 and 128, respectively. For cross-domain retrieval, the training samples from the source domain are used as retrieval database. For the single-domain retrieval, the training samples from the target domain are used as retrieval database.

To further prove our versatility and cross-domain retrieval performance, a large number of experiments were carried out on \emph{Office-Home}. For the sake of simplicity,  Artistic images, Clip Art images, Product images, and Real-World images are replaced as A, C, P and R, respectively. A$\rightarrow$C implies Artistic is the source domain and Clip Art is the target domain.

\subsection{Experimental Results}
In Table \ref{tab1}, we report the MAP scores (\%) of all the compared methods and our method PWCF on \emph{MNIST\&USPS}, \emph{VOC2007\&Caltech101}, and \emph{Caltech256\&ImageNet} for cross-domain retrieval. Obviously, our PWCF outperforms compared methods on all databases in most cases. To further prove the effectiveness of our method, we conduct an experimental evaluation of single-domain retrieval. The results are shown in Table \ref{tab2}. We can see that our method is superior to the compared methods in both cross-domain retrieval and single-domain retrieval.
\begin{table}
\centering
\caption{The MAP scores (\%) on Office-home databases with 64 bits for cross-domain retrieval. P$\rightarrow$R implies Product is the source domain and Real-World is the target domain.} \begin{comment}{Artistic images (\textbf{A}), Clip Art images (\textbf{C}), Product images (\textbf{P}) and Real-World images (\textbf{R})} \end{comment}
%\small
\resizebox{84mm}{21.35mm}{
\begin{tabular}{c|cccccc|c}
\hline
  &P$\rightarrow$R&R$\rightarrow$P&C$\rightarrow$R&R$\rightarrow$C&A$\rightarrow$R&R$\rightarrow$A&average \\
  \hline
 NoTL&25.02&29.09&16.98&14.73&27.38&21.05&22.38\\
SH&15.01&16.82&10.38&8.32&14.52&12.41&12.91\\
ITQ&26.32&29.13&17.60&15.88&26.86&21.99&22.96\\
SDH&8.86&7.72&6.44&6.15&10.07&9.11&8.06\\
LSH&10.74&14.78&9.85&8.37&12.24&10.02&11.00\\
SGH&24.12&26.47&16.10&14.14&22.82&20.41&20.68\\
OCH&20.06&20.23&10.97&10.61&19.35&15.32&16.09\\
ITQ+&17.94&--&10.61&--&--&15.16&14.57\\
LapITQ+&15.94&--&11.72&--&--&13.52&13.83\\
GTH&19.40&22.80&12.27&12.03&20.98&16.47&17.33\\
KSH&32.02&34.42&21.56&\textbf{18.51}&25.87&20.04&25.40\\
SDH&25.75&27.90&15.97&16.72&32.06&22.79&23.53\\
\hline
PWCF&\textbf{34.03}&\textbf{34.44}&\textbf{24.22}&18.42&\textbf{34.57}&\textbf{28.95}&\textbf{29.11}\\
\hline
\end{tabular}}
\label{tab3}
\end{table}

\begin{figure}
 \centering
  \begin{minipage}{4cm}
 \centerline{\includegraphics[scale=0.27]{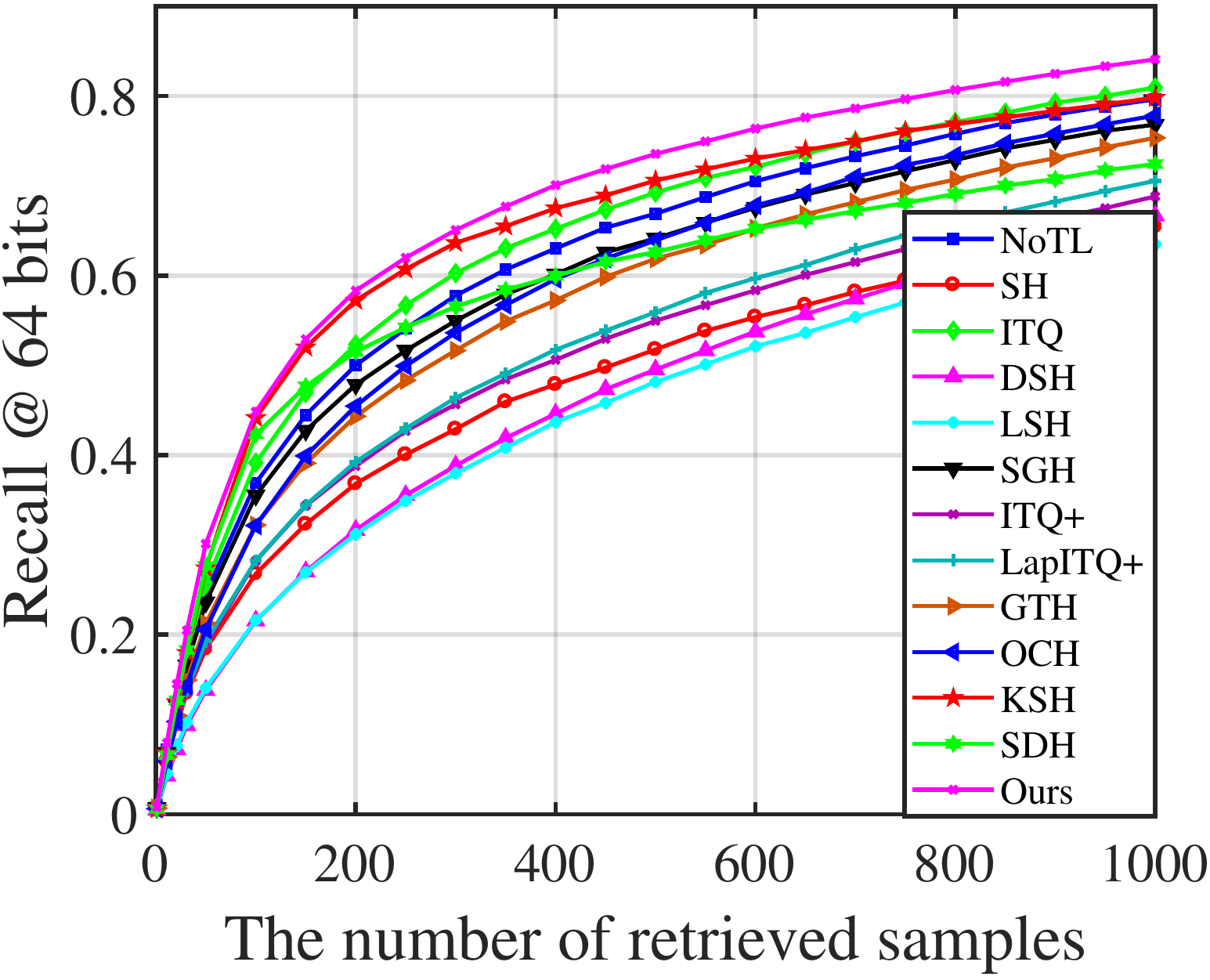}}
 \centerline{(a)}
\end{minipage}
\hfill
\begin{minipage}{4cm}
 \centerline{\includegraphics[scale=0.27]{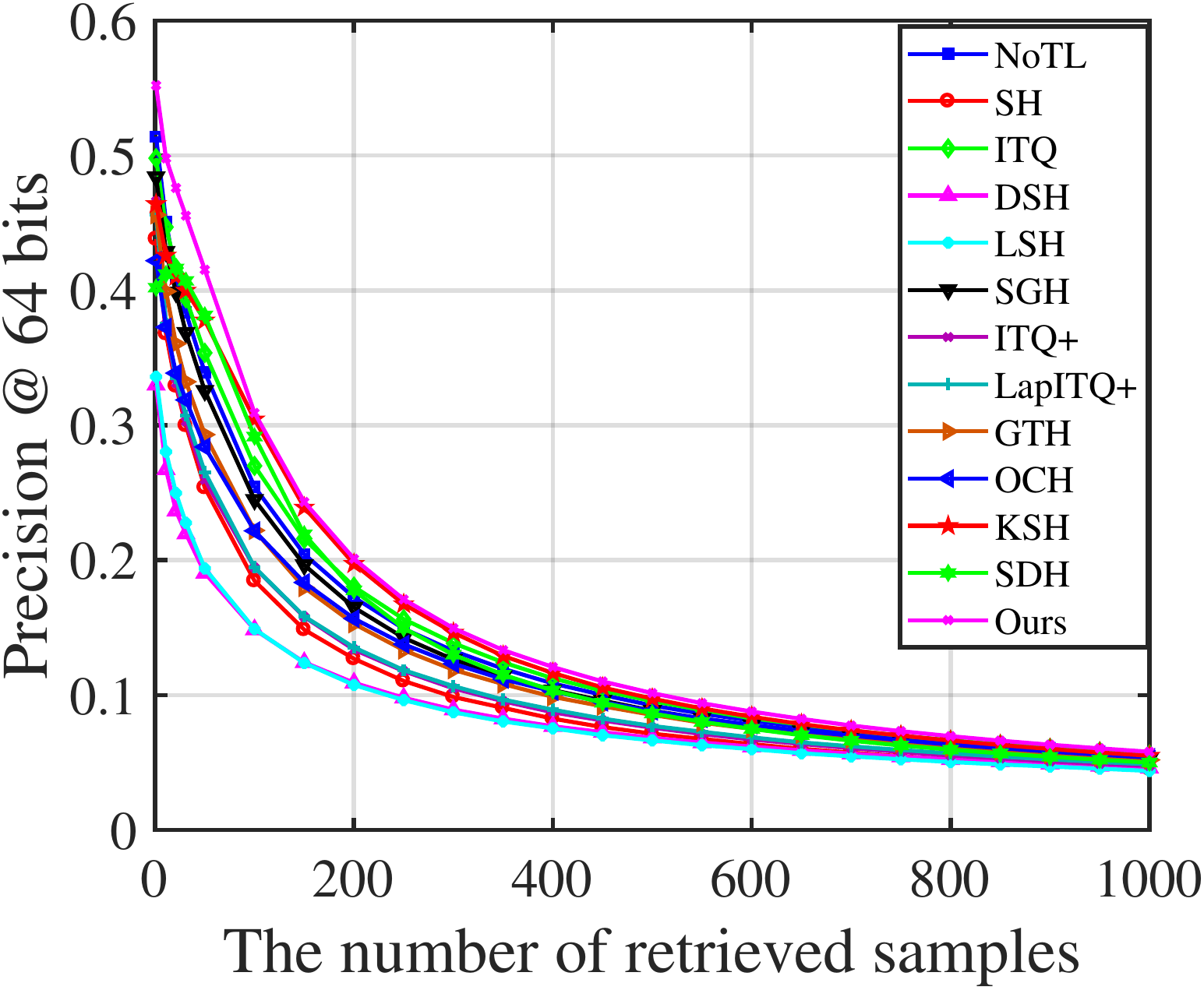}}
 \centerline{(b)}
\end{minipage}
\hfill
  \caption{The influence of the number of retrieved samples for cross-domain retrieval on Product$\rightarrow$Real with 64 bits.}
  \label{fig5}
\end{figure}

\begin{table*}
\centering
\caption{The MAP scores (\%) on MNIST\&USPS databases with varying code length from 16 to 128 for cross-domain retrieval or single-domain retrieval.}
\small
\resizebox{165mm}{18.2mm}{
\begin{tabular}{c|cccccc|cccccc}
\hline
        &  \multicolumn{6}{c|}{cross-domain}  & \multicolumn{6}{c}{single-domain}    \\
\hline
 Bit    &16&32&48&64&96&128     &16&32&48&64&96&128   \\
\hline
PWCF  & \textbf{47.47}& \textbf{51.99}&\textbf{51.44} &\textbf{51.75}&\textbf{50.89}&\textbf{53.95}& \textbf{69.37}& \textbf{70.70}&\textbf{70.94} &\textbf{71.64}&\textbf{73.51}&\textbf{73.89}\\
PWCF-T&45.13&48.76&50.02&50.57&50.66&51.43&51.45&62.18&68.25&69.91&71.62&72.81\\
PWCF-F&44.07&45.40&47.58&50.01&50.12&53.18&52.55&62.70&66.37&69.07&71.27&72.03\\
PWCF-M&29.13&30.76&33.44&34.56&35.68&36.22&62.16&67.84&69.04&70.25&70.68&71.32\\
PWCF-C&40.80&47.02&50.04&50.36&48.88&47.81&50.85&61.11&64.94&66.40&69.24&70.30\\
PWCF-H&29.09&32.14&34.89&36.36&36.84&36.88&49.12&64.13&67.82&69.15&70.64&71.09\\
PWCF-Q&10.89&10.65&10.64&10.60&10.63&10.52&15.22&15.43&16.24&16.83&19.05&18.49\\
\hline
\end{tabular}}
\label{tab4}
\end{table*}
In Table \ref{tab3}, we report the MAP scores (\%) of all the compared methods and our PWCF with 64 bits on \emph{Office-Home} for cross-domain retrieval. We can see however the source domain and target domains are set, our methods perform better than others. The results certify that our PWCF has universality in practical application. We also show the influence of the number of retrieved samples for cross-domain retrieval task. Fig. \ref{fig5} (a) shows the precision when the number of retrieved samples vary from 0 to 1000 and Fig. \ref{fig5} (b) shows the recall when the number of retrieved samples vary from 0 to 1000. From the figures, we can see that our PWCF always presents competitive retrieval performance compared to baselines, which demonstrates the efficacy of our PWCF.

\subsection{Convergence Analysis}
\begin{figure}[h]
  \centering
  \begin{minipage}{4cm}
 \centerline{\includegraphics[scale=0.27]{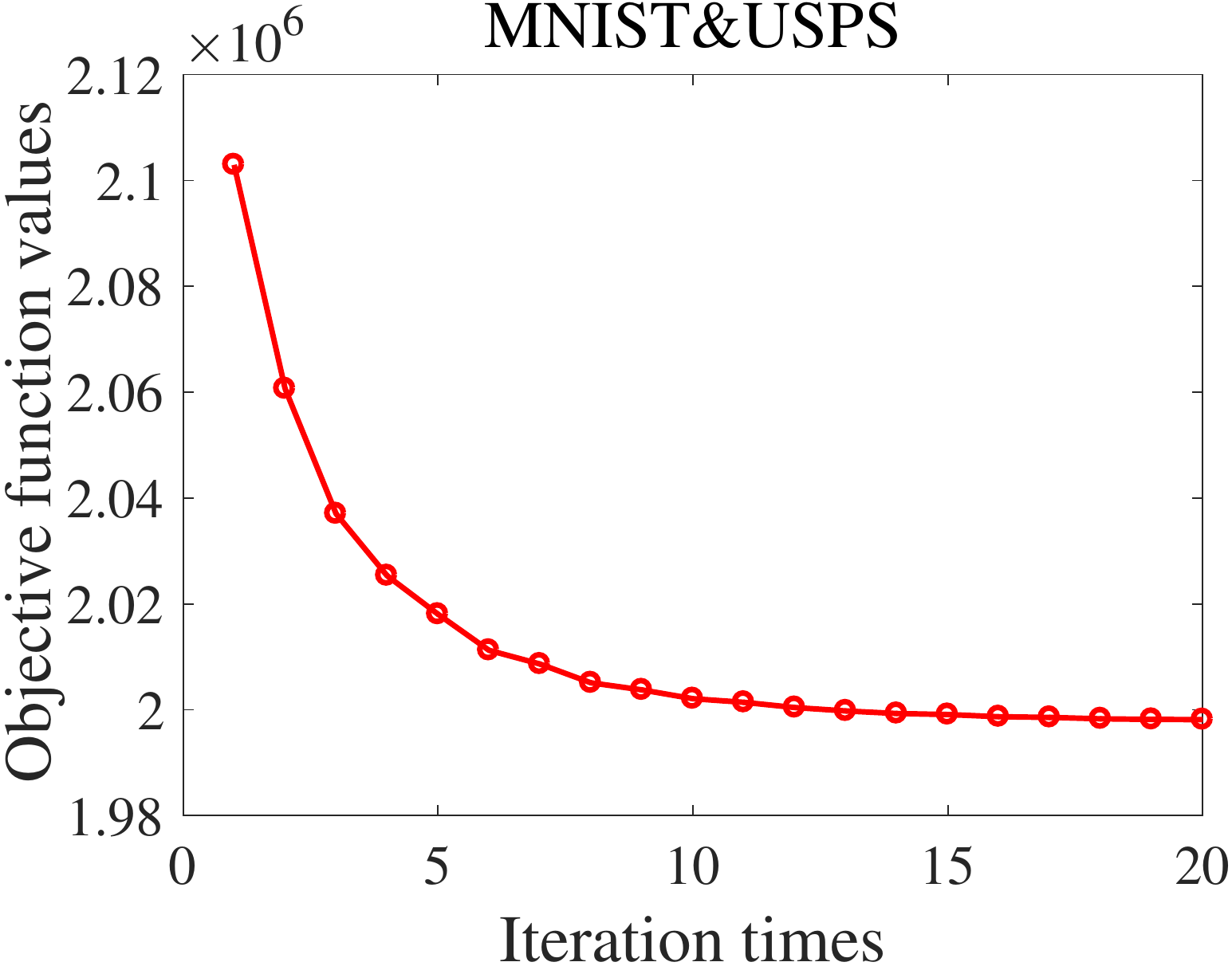}}
 \centerline{(a)}
\end{minipage}
\hfill
\begin{minipage}{4cm}
 \centerline{\includegraphics[scale=0.27]{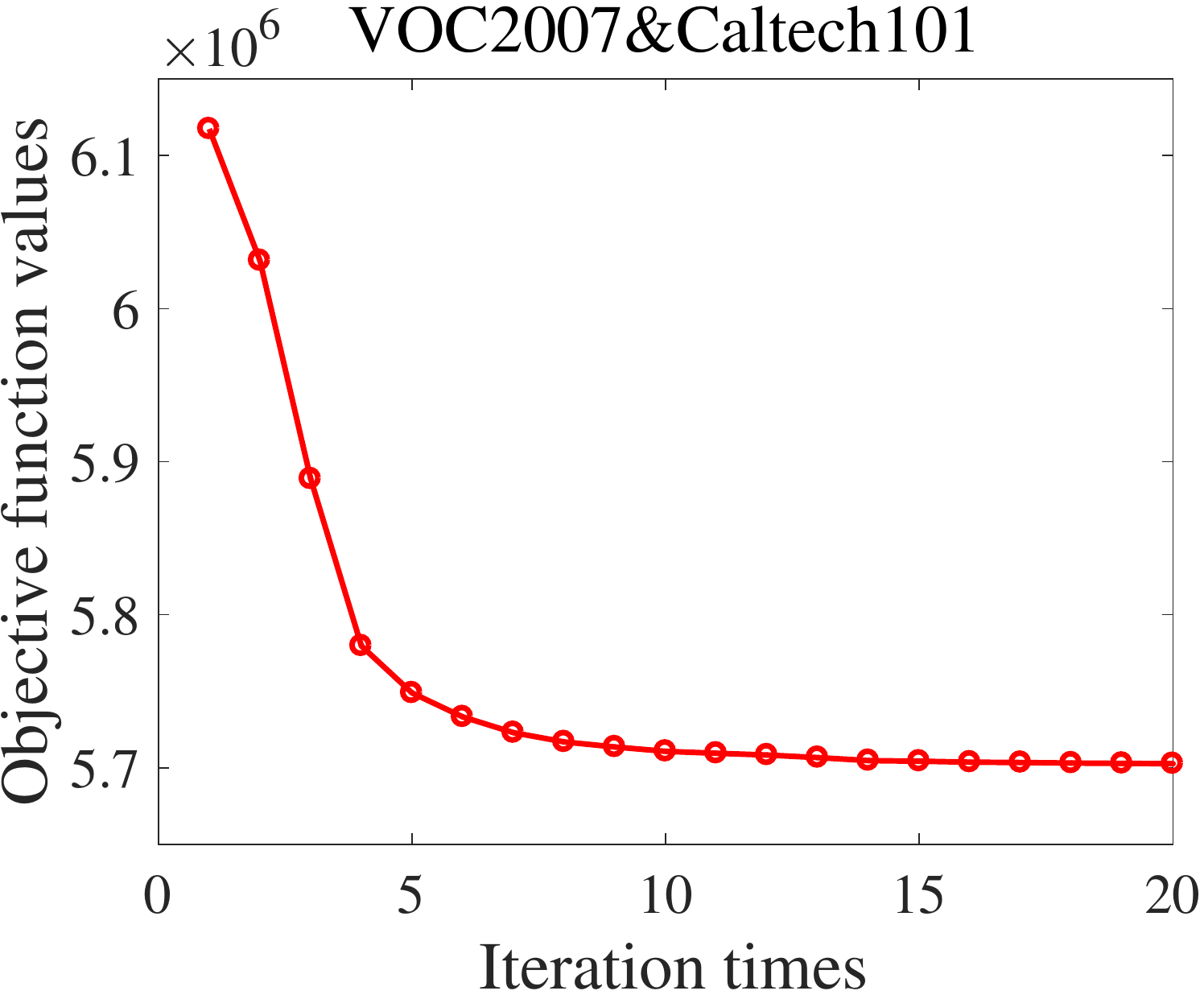}}
 \centerline{(b)}
\end{minipage}
  \caption{Convergence of Algorithm 1 on (a) MNIST\&USPS and (b)VOC2007\&Caltech101 with 64 bits. }
  \label{fig6}
\end{figure}

The proposed PWCF is solved with a variable alternating strategy, and the convergence can be guaranteed. We present
the convergence curves of the objective function in Fig. \ref{fig6}, from which we see that PWCF can quickly converge to an optimal solution within several iterations.
\subsection{Ablation Study}
We investigate six variants of PWCF in Table \ref{tab4}: (1) PWCF-T is the PWCF variant without BP induced focal-triplet loss $\mathcal{T}ri$. (2) PWCF-F is the STH variant, which replaces the BP induced focal-triplet loss as the standard triplet loss. (3) PWCF-M is the PWCF variant without manifold loss $\mathcal{M}$. (4) PWCF-C is the PWCF variant without BP induced classifier loss $\mathcal{C}$. (5) PWCF-H is the PWCF variant without the Histogram Feature of Neighbors, which calculates the weight matrix $\mathbf{Z}$ by using original content features. Also, the hard triplets are constructed by original features without the Histogram Feature of Neighbors. (6) PWCF-Q is the PWCF variant without BP induced quantization loss $\mathcal{Q}$. We report the results to different code lengths on the \emph{MNIST\&USPS} dataset for single-domain and cross-domain retrieval.

We can see that the four parts of our model have different effects on retrieval performance. Comparing PWCF-T, PWCF-F, PWCF, triplet loss is good for training PWCF and our proposed focal-triplet hashing loss is better than standard triplet loss. Comparing PWCF-M with PWCF, the underlying manifold structure across different domains is extremely helpful to capture the correlation between samples. Comparing PWCF-H with PWCF, the proposed histogram features reduce the impact of data distribution discrepancy between different domains and it is unreasonable to use Euclidean distance of original features to measure the similarity between cross-domain samples.

\section{Conclusion}
In this paper, we propose an effective domain adaptive retrieval method named Probability Weighted Compact Feature Learning (PWCF), which learns compact binary codes to represent images. First, we propose BP induced focal-triplet loss, BP induced quantization loss and BP induced classification loss from the Bayesian perspective to optimize the binary compact feature between samples from different domains. Besides, The underlying manifold structure across different domains is used to capture meaningful nearest neighbors of different domains and further explore the potential correlation. To address the data distribution discrepancy issue, we propose a Histogram Feature of Neighbors (HFON) to metric the similarity/dissimilarity between the samples from different domains. The experimental results show that our PWCF always shows much higher retrieval performance in the scenarios of the cross-domain retrieval and single-domain retrieval, which verify that our method outperforms many state-of-the-art image retrieval methods.

\textbf{Acknowledgement:} This work was supported by the National Science Fund of China under Grants (61771079),
Chongqing Youth Talent Program, and the Fundamental Research Funds of Chongqing (No. cstc2018jcyjAX0250).
{\small
\bibliographystyle{ieee_fullname}
\bibliography{egbib}
}

\end{document}